\documentclass[10pt,twocolumn,letterpaper]{article}
\usepackage{cvpr}              
\usepackage{cuted}

\makeatletter
\newcommand*{\rom}[1]{\expandafter\@slowromancap\romannumeral #1@}
\makeatother

\definecolor{cvprblue}{rgb}{0.21,0.49,0.74}
\usepackage[pagebackref,breaklinks,colorlinks,citecolor=cvprblue]{hyperref}

\def\paperID{8058} 
\def\confName{CVPR}
\def\confYear{2025}

\title{SerialGen: Personalized Image Generation by First Standardization Then Personalization}

\author{
Cong Xie${^*}$\qquad
Han Zou${^*}$\qquad
Ruiqi Yu\qquad
Yan Zhang${^{\dagger}}$\qquad
Zhenpeng Zhan${^{\dagger}}$\qquad\\
Global Business Unit, Baidu Inc.
\\ 
{\color{red} \url{https://serialgen.github.io}}
}

\newcommand\extrafootertext[2]{%
    \bgroup
    \renewcommand\thefootnote{\fnsymbol{footnote}}%
    \renewcommand\thempfootnote{\fnsymbol{mpfootnote}}%
    \footnotetext[#1]{#2}%
    \egroup
}

\begin{document}
\maketitle
\extrafootertext{1}{Equal contributions. $^{\dagger}$Corresponding authors.}

\begin{abstract}
In this work, we are interested in achieving both high text controllability and whole-body appearance consistency in the generation of personalized human characters.
We propose a novel framework, named SerialGen, which is a serial generation method consisting of two stages: first, a standardization stage that standardizes reference images, and then a personalized generation stage based on the standardized reference. 
Furthermore, we introduce two modules aimed at enhancing the standardization process.
Our experimental results validate the proposed framework's ability to produce personalized images that faithfully recover the reference image's whole-body appearance while accurately responding to a wide range of text prompts. Through thorough analysis, we highlight the critical contribution of the proposed serial generation method and standardization model, evidencing enhancements in appearance consistency between reference and output images and across serial outputs generated from diverse text prompts. The term "Serial" in this work carries a double meaning: it refers to the two-stage method and also underlines our ability to generate serial images with consistent appearance throughout.
\end{abstract}    
\section{Introduction}
\label{sec:intro}

Recently, text-to-image generation models based on diffusion methods~\cite{ramesh2021zero,saharia2022photorealistic,rombach2022high,ramesh2022hierarchical,podell2023sdxl} have experienced rapid development. These models demonstrate a remarkable capability to control the generated content based on the text prompt. Concurrently, the personalized text-to-image generation tasks~\cite{ye2023ip,wang2024instantid,guo2024pulid,xiao2024fastcomposer,li2024photomaker,ruiz2023dreambooth,gal2022image,zhou2024storymaker,chen2023subject,arar2023domain,gal2024lcm} have garnered widespread attention due to their broad range of application scenarios. Given a reference image of a subject, the objective is to output additional images of the subject based on text prompts with high text controllability and appearance consistency. Text controllability demands a close match between the generated images and the text prompt. Appearance consistency ensures consistency between the generated images and the input subject. One category of work requires fine-tuning~\cite{ruiz2023dreambooth,gal2022image,kumari2023multi} the model for each new input subject. The practical application of this type of method is limited due to the high cost of fine-tuning. 

In this work, we focus on tuning-free methods for personalized generation of human characters~\cite{ye2023ip,xiao2024fastcomposer,zhou2024storymaker} , ensuring text controllability and \textit{whole-body appearance}, including facial features, hairstyles, and clothing.
To exempt from fine-tuning for every new subject, another type of approach typically involves a training phase on a substantial amount of subject data. The trained model is expected to generalize to new subjects, thereby avoiding the need for fine-tuning during the inference stage. Despite being more practical, this approach faces significant challenges. The training pairs of input (reference) and output (target) images utilized in existing methods typically derive from the same source image, generated by simple data augmentation operations such as cropping and flipping. Such training data poses significant challenges to model training due to the highly consistent content between the input and output images. 

In this study, we empirically observe that models trained with such data do not effectively achieve the goal of personalized generation tasks: they either compromise text controllability to maintain high appearance consistency or sacrifice appearance consistency to enhance text controllability. If the model is sufficiently powerful, it can easily replicate the reference image in the output to achieve minimal loss, thus achieving high appearance consistency but displaying inadequate responsiveness to text prompts such as an action different from the reference. Conversely, if the model is constrained to extract only compressed information from the reference, it often fails to capture all the appearance-related visual features and discards irrelevant features. This leads to a model that better aligns with text prompts, albeit at the cost of compromised appearance consistency.

To alleviate the above-mentioned problems, we propose to train personalized models using paired images of (standardized reference image, target image), where both images share the same appearance but differ in non-appearance elements (NAE), such as background, pose, expression, and viewpoint. The standardized reference image will possess a standardized NAE, whereas the target image will feature non-standard NAEs. In addition to addressing the replication issue, another motivation for using standardized reference is that they simplify the generation task. It is easier for a feature extractor to process standardized input than to handle inputs with complex and varied characteristics. By standardizing NAE of the input images, the model can focus more on capturing the appearance features, leading to better personalized character generation. 

To achieve this goal, we introduce SerialGen, which is a serial generation method consisting of two stages: first, a standardization stage that standardizes reference images, and then a personalized generation stage based on the standardized reference. The advantages of this framework are threefold:
\begin{itemize}
\item Employing different images as reference-target pairs can mitigate the issue of input-output replication, even when utilizing a powerful model. 
\item  Utilization of standardized reference enhances appearance consistency between reference image and output image. 
\item  Standardization enhances appearance consistency across serial images generated from different text prompts, which is highly beneficial in practical applications, such as comic story generation. For instance, if a reference image displays only a character's head, without standardization, the body's appearance could vary in different generation processes. By standardizing, the body is pre-generated in the standardized reference image, ensuring consistent appearance across various text prompts.
\end{itemize}

Our contributions can be summarized as follows: 1) We introduce a novel first-standardization-then-personalization framework enabling to generation of images with high text controllability and maintaining superior whole-body appearance consistency.
2) Our comprehensive analysis underlines the pivotal role of the serial generation method and the standardization model, highlighting several key benefits.
3) We propose two innovative modules designed to augment the standardization model. The effectiveness of these modules is validated through experimental results.
\section{Related Works}

Personalized text-to-image generation based on diffusion models has experienced rapid development. These models can be categorized into two primary types: methods that require subject-specific fine-tuning and those that don't. Models necessitating fine-tuning, such as those proposed by ~\cite{gal2023encoder,kumari2023multi,gal2022image,ruiz2023dreambooth}, demand additional training for each new input subject, which may limit their practicality. Conversely, models that forego fine-tuning offer enhanced flexibility, eliminating the need for further training. 
The bulk of research in this domain has concentrated on preserving facial identity in generated images by conditioning on facial images, as noted in works by ~\cite{wang2024instantid,ye2023ip,han2025face,he2024imagine,li2024photomaker,guo2024pulid,gal2024lcm}. 
The facial features are extracted by pre-trained face recognition models~\cite{deng2019arcface,deng2020subcenter} and then integrated into diffusion models. Recently, efforts~\cite{ye2023ip,xiao2024fastcomposer,zhou2024storymaker} have been directed towards generating images that maintain the full body appearance of characters, encompassing facial identity, clothing, hairstyle, and other attributes. 

Many recent methods train with unpaired images, leading to significant training difficulties and causing models to replicate these images. Efforts to collect paired human face data~\cite{li2024photomaker,he2024imagine} have involved generating and filtering images through facial recognition for similarity. However, this struggles with maintaining consistency in whole-body appearance, including clothing and hairstyles, due to difficulties in varying poses, backgrounds, and perspectives. Additionally, there’s no model that effectively filters data based on whole-body appearance.

Recent developments have made significant strides in generating multiple images or videos from a single reference image of a character using diffusion models~\cite{wang2024disco,xu2024magicanimate,hu2024animate,chang2023magicpose,song2020denoising,zhong2025posecrafter,hu2022lora,kim2024tcan,zhu2024champ}. These models are adept at capturing all details from the reference image, maintaining both the whole-body appearance of character and the background scene unchanged across a sequence of generated images. 

Furthermore, StoryDiffusion~\cite{zhou2024storydiffusion} proposes a consistent self-attention mechanism to preserve character consistency across a sequence of generated images.  However, maintaining appearance consistency between the reference and target images is not the primary focus of their work.
\section{Methods}
\label{sec:method}

\subsection{Preliminaries}
\paragraph{Personalized Text-to-Image Generation Models}
The widely adopted framework for personalized text-to-image generation typically comprises two main components: a diffusion model and a reference encoder. The reference encoder is tasked with extracting visual features from the reference image, which are then integrated into the diffusion model. Latent diffusion models~\cite{rombach2022high,podell2023sdxl} are frequently utilized within this context. A training image $\mathbf{x}$ is first converted into a latent representation through the VAE~\cite{esser2021taming} encoder $\mathbf{z} = \varepsilon(\mathbf{x})$. Subsequently, a noise $\epsilon$ is imposed to $\mathbf{z}$ at timestep $t$, resulting in a noised latent $\mathbf{z}_t$. A denoising UNet $\epsilon_{\theta}$ is employed to estimate the imposed noise. The training loss is defined as follows:
\begin{align}
\label{eq:diff}
\mathcal{L}=\mathbb{E}_{\mathbf{x},c,\epsilon,t}\left\|\epsilon-\epsilon_{\theta}(\mathbf{z}_t,t,c,\phi_{\eta}(\mathbf{x}))\right\|_2^2
\end{align}
where $c$ refers to the text embeddings extracted from text prompts corresponding to $\mathbf{x}$ and $\phi_{\eta}$ is the reference encoder. 

\paragraph{Human Image Animation Models} 
Our proposed framework is also closely related to a diffusion-based model for human image animation~\cite{hu2024animate}. This model employs a dual-stream structure with a Denoising UNet and a ReferenceNet, which share the same backbone architecture. The Denoising UNet processes a guided pose alongside a noised image to predict the noise, whereas the ReferenceNet is responsible for encoding the reference image and integrating this information into the Denoising UNet. Training an animation model requires paired images of reference-target: $(\mathbf{x}^r, \mathbf{x}^o)$.
The training loss is defined as follows:
\begin{align}
\mathcal{L}=\mathbb{E}_{(\mathbf{x}^r, \mathbf{x}^o),c,\epsilon,t}\left\|\epsilon-\epsilon_{\theta}(\mathbf{z}^o_t,t,c,\phi_{\delta}(\mathbf{x}^r))\right\|_2^2
\end{align}
where $\mathbf{z}^o_t$ is $\mathbf{z}^o = \varepsilon(\mathbf{x}^o)$ diffused with noise $\epsilon$ at timestep $t$, $c$ is the guide poses, $\phi_{\delta}$ denotes the ReferenceNet.
The integration of ReferenceNet features is detailed as follows: For a given self-attention module within the ReferenceNet, the input feature map $\mathbf{f}_r \in \mathbb{R}^\mathrm{h \times w\times c}$ is concatenated to the input feature map $\mathbf{f}_d \in \mathbb{R}^\mathrm{h\times w\times c}$ of the identical self-attention module in the Denoising UNet. This concatenation occurs along the spatial dimension to produce $\mathbf{f}_c \in \mathbb{R}^\mathrm{2h\times w\times c}$. Subsequently, this concatenated feature map $\mathbf{f}_c$ is input into the self-attention module of the Denoising UNet. For further details, refer to the original paper.

\begin{figure*}[!ht]
  \centering
   \includegraphics[width=1\textwidth]{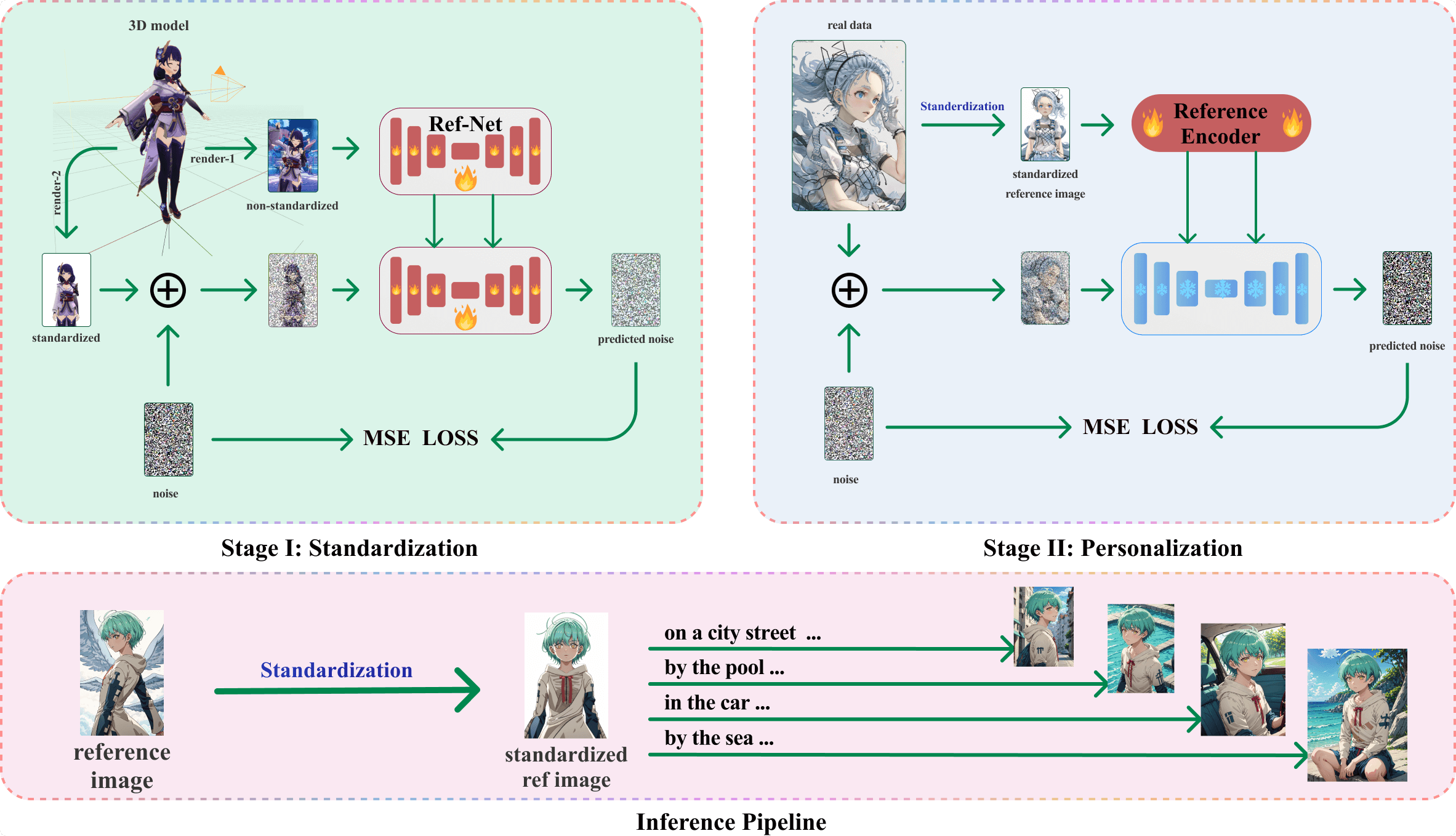}
   \caption{Overview of the proposed SerialGen with two stages: (1) \textit{Standardization} – training a standardization model on synthetic data, and (2) \textit{Personalization} – using the standardization model to create (standardized reference, target) pairs for personalized text-to-image model training. During inference, once a reference image is standardized, serial images can be generated based on different text prompts.}
   \label{fig:overview}
\end{figure*}
\subsection{Overall Framework}

The overall framework of the proposed SerialGen is illustrated in Figure \ref{fig:overview}. SerialGen comprises two stages. In the first stage, a standardization model is introduced and trained on synthetic data. In the second stage, this standardization model is utilized to convert a set of real training images to standardized images. This process results in pairs of (standardized reference image, target image), which are then used to train the personalized text-to-image model. The inference pipeline is also divided into two steps. Upon receiving a reference image, it is first standardized, followed by the personalized text-to-image generation.

\begin{figure}[t]
  \centering
   \includegraphics[width=1\linewidth]{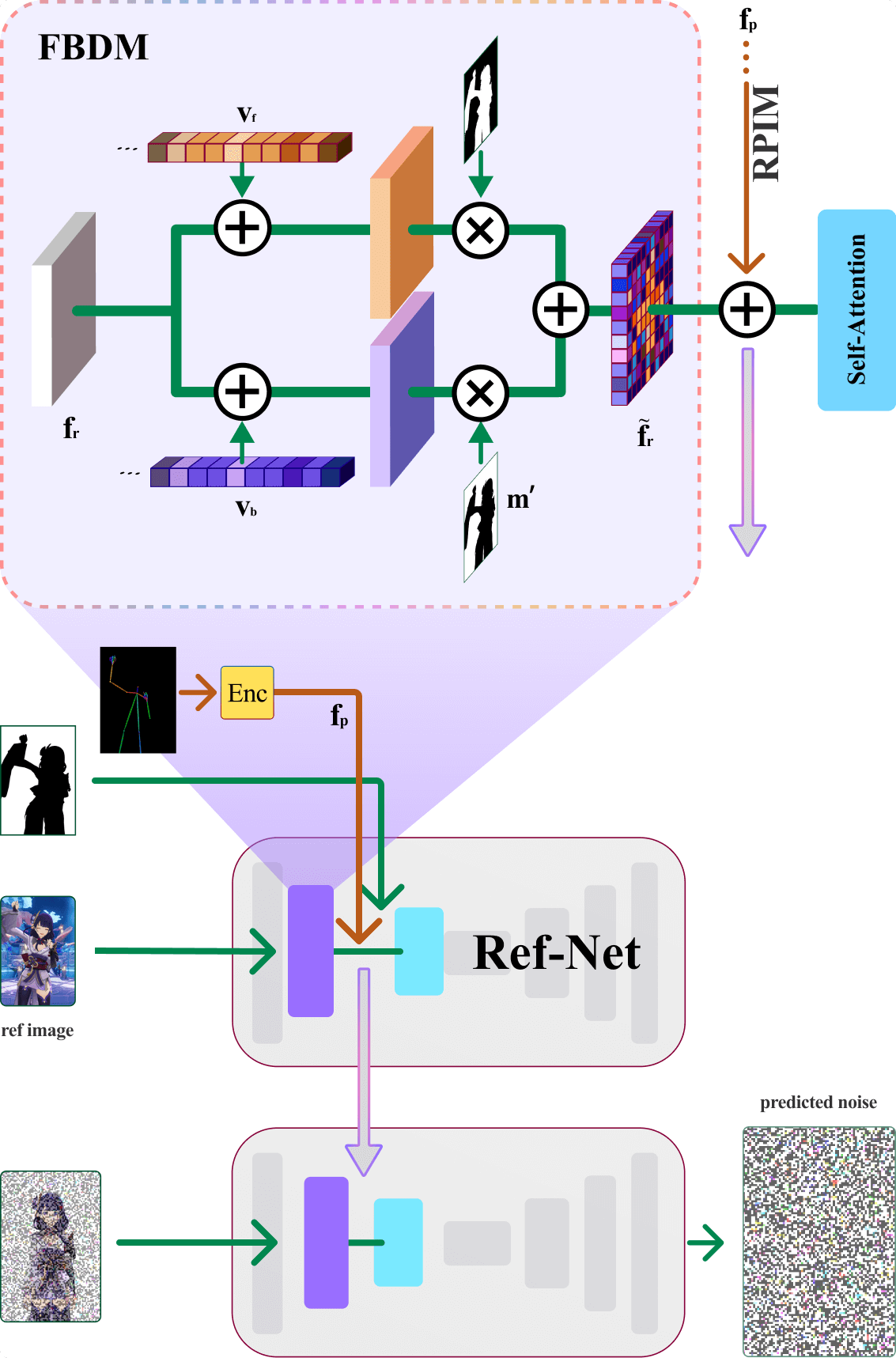}
   \caption{Illustration of the standardization model. The pose and mask of the reference image are input into ReferenceNet to enhance the effect.}
   \label{fig:stage1}
\end{figure}

\subsection{Stage ~\rom{1} : Standardization}
\label{sec:standardization}

Given any reference image, Stage ~\rom{1} transforms it into a new image with standardized NAE, while preserving the referenced appearance unchanged. To accomplish this, we introduce a standardization model.

\paragraph{Standardization Model}
Inspired by the tasks of human image animation, which also involves altering human poses while maintaining the appearance invariant, we use a state-of-the-art human image animation framework~\cite{hu2024animate} without temporal attention module as our baseline model.
To enhance the standardization of the background and pose, we propose two modules designed to improve the baseline model: 1) a Foreground-Background Distinction Module (FBDM), which is intended to explicitly integrate the background mask of the reference image into the model, and 2) a Reference Pose Injection Module (RPIM), aimed at explicitly incorporating the pose information from the reference image into the model, thereby enabling the model to accurately locate body parts. An overview of the standardization model is given in Figure \ref{fig:stage1}.
\paragraph{Foreground-Background Distinction Module}
Within each self-attention module of the ReferenceNet, we introduce two learnable class tokens: foreground token $\mathbf{v}_f \in \mathbb{R}^{1\times 1\times c}$ and background token $\mathbf{v}_b\in \mathbb{R}^{1\times 1\times c}$. Given a background mask $\mathbf{m}_b\in \mathbb{R}^{h\times w}$, $\mathbf{v}_f$ and $\mathbf{v}_b$ are separately added to the foreground and background regions of input feature map $\mathbf{f}_r \in \mathbb{R}^{h\times w\times c}$ in accordance with the mask:
\begin{equation}
  \tilde{\mathbf{f}_r} = (\mathbf{f}_r + \mathbf{v}_{b})\circ \mathbf{m}' + (\mathbf{f}_r + \mathbf{v}_{f})\circ (1 - \mathbf{m}')
  \label{eq:fgbg}
\end{equation}
where $\mathbf{m}'\in \mathbb{R}^{h\times w}$ is resized version of $\mathbf{m}$ to fit the resolution of $\mathbf{f}_r$.
This method enables the explicit incorporation of mask information into the self-attention mechanism.
 
\paragraph{Reference Pose Injection Module}
This module integrates pose information into the model. Specifically, the pose of the reference image is extracted utilizing DWPose~\cite{yang2023effective}. Following this, a pose feature map is derived from the pose image via a light convolutional network composed of four layers. Within each self-attention module of the ReferenceNet, an additional convolutional layer is utilized to adjust the channel number of the pose feature to match the self-attention module. More details are given in the supplementary material. Furthermore, the resolution of the pose feature is interpolated to match that of the self-attention module. This processed pose feature $\mathbf{f}_p$ is subsequently added to $\tilde{\mathbf{f}_r}$:
\begin{equation}
\begin{aligned}
\tilde{\mathbf{f}_r} = \tilde{\mathbf{f}_r} + \mathbf{f}_p
\end{aligned}
\end{equation}
$\tilde{\mathbf{f}_r}$ is then concatenated to the feature $\mathbf{f}_d$ of Denoising U-Net for further processing.

\paragraph{Synthetic Data}
To train the standardization model, paired images consisting of (non-standardized, standardized) are essential. However, the collection of such real-world data is expensive. In response to this challenge, we create a synthetic dataset by rendering animatable 3D character models. As illustrated in Figure \ref{fig:overview}, given a 3D character model, images of the character in various poses, backgrounds, expressions, and viewpoints are rendered to represent the non-standardized images. Concurrently, utilizing the same character model, the standardized image is rendered adhering to a standard pose, against a white background, with a neutral expression, and the face scaled to a fixed position. This rendering pipeline was automated to generate a substantial volume of paired images for training purposes.

\subsection{Stage ~\rom{2}: Personalization}
\label{sec:personalization}
At this stage, pairs of images, consisting of (standardized reference image, target image), can be constructed utilizing a standardization model. As illustrated in Figure \ref{fig:overview}, a real image is standardized using a standardization model. The standardized output acts as the standardized reference image and the original real image serves as the target image. By applying standardization to a substantial corpus of real images, we can generate numerous paired images, which are crucial for training a personalized model. 

One might question the domain gap problem that arises when the standardization model, trained on synthetic data, is applied to real data for inference. Our experimental results demonstrate that while the appearance of the output image remains unchanged, its style is indeed biased towards the 3D style used during training. This is exemplified in the supplementary material. However, this bias does not impede the training or inference of personalized models. The reason is that the biased 3D style can be considered an integral part of the standardization. This bias can be effectively mitigated after training with real target images.

Given paired data, a personalized model can be trained using the loss function described in Equation \eqref{eq:diff}, with modifications applied to the input of the reference encoder:
\begin{align}
\label{eq:diff2}
\mathcal{L}=\mathbb{E}_{\mathbf{x},c,\epsilon,t}\left\|\epsilon-\epsilon_{\theta}(\mathbf{z}_t,t,c,\phi_{\eta}(\varphi(\mathbf{x}))\right\|_2^2
\end{align}
where $\varphi$ is the standardization model frozen during the training of the second stage.
\section{Experiments}

\subsection{Implementation Details}
\paragraph{The Standardization Stage} For the creation of synthetic data (non-standardized, standardized), we collect 3D character models sourced from publicly accessible websites. We filtered out characters that were not animatable, exhibited duplicate shapes, lacked skeletal structures, or presented texture defects. Following this filtration process, a total of 2,924 3D character models were retained. These models were subsequently paired with 18 distinct motion models, encapsulating 3,817 unique motion poses. The ensuing combination resulted in the generation of 10,513,090 data pairs, each rendered with a transparent background. 
During the training of the standardization model, non-standardized images were dynamically merged with 9,186 background images. The training was conducted utilizing 8 NVIDIA A100 GPUs, with a batch size of 4 per GPU. The training was conducted at a resolution of $512\times768$ for a duration of 50,000 steps. During the inference phase, reference images are padded to  $512\times768$ and generated utilizing the DDIM~\cite{song2020denoising} sampler, which employs 20 denoising steps. 

\paragraph{The Personalization Stage} 
We adopt the existing model architecture, IP-Adapter~\cite{ye2023ip}, as the implementation for our second-stage personalization model. Specifically, we utilize Stable Diffusion XL (SDXL)~\cite{podell2023sdxl} as our diffusion model, OpenCLIP ViT-H/14~\cite{radford2021learning} as the reference encoder, and a multilayer perceptron (MLP) layer for the adapter module, which generates 257 token features for cross-attention. The training dataset comprises approximately 300,000 character images, all of which have been standardized by trained standardization model, resulting in around 300,000 image pairs. During the training process, the adapter modules undergo training, while the rest part is kept frozen. The training procedure is conducted at a resolution of $512\times768$, spanning 200,000 steps on 8 NVIDIA A100 GPUs, with a batch size of 8 per GPU. During the inference phase, we employ the DDIM sampler with 20 denoising steps. 

\subsection{Quantitative and Qualitative Results}
\label{sec:quantitative}

We compare our method with three peer personalized models, all of which are tuning-free and capable of utilizing a full-body reference image. These models include StoryMaker~\cite{zhou2024storymaker}, IP-Adapter~\cite{ye2023ip}, and FastComposer~\cite{xiao2024fastcomposer}. Following previous practices~\cite{ye2023ip}, 
we utilize the CLIP image similarity score (CLIP-I) to assess appearance consistency and the CLIP text similarity score (CLIP-T) to evaluate text controllability.
To more precisely measure the similarity between two character images, we first remove the background regions of the images before computing the CLIP-I scores. 
Additionally, we employ face similarity (Face Sim.), a metric quantifying the cosine similarity between identity embeddings extracted using ArcFace~\cite{deng2019arcface}.
To evaluate performance, we compiled a test dataset consisting of 40 characters, which includes 20 anime characters and 20 real-life individuals. We generated 20 distinct text prompts using ChatGPT-4, encompassing a diverse range of actions, backgrounds, viewpoints, and expressions. For each combination of prompt and character, we produced four images with randomness introduced. Unless specified otherwise, all subsequent experiments were conducted on this test dataset.

\begin{figure*}[ht]
  \centering
   \includegraphics[width=0.9\textwidth]{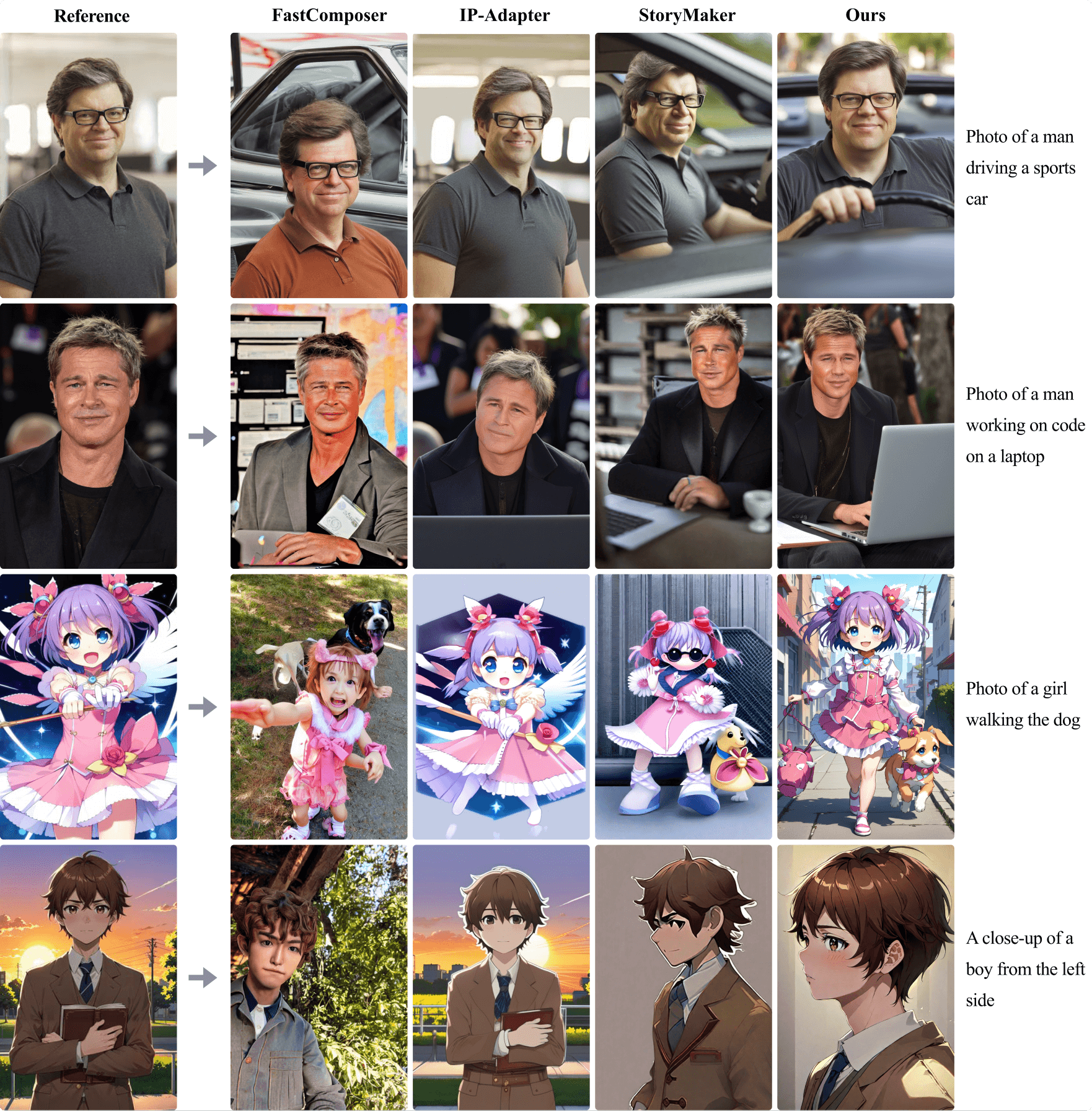}
   \caption{Comparison with other methods. Our method is capable of generating images with high text controllability and appearance consistency.
}
   \label{fig:qualitative_comp}
\end{figure*}

As presented in Table~\ref{tab:comp_clip_score}, our method achieves the highest scores for both CLIP-T and CLIP-I. These results suggest that our proposed method surpasses others in terms of appearance consistency and text controllability.
For the Face Sim. metric, our method ranks second, following StoryMaker. Notably, StoryMaker employs an additional face encoder to enhance facial consistency.
Qualitative evidence supporting our findings is illustrated in Figure~\ref{fig:qualitative_comp}, where it is evident that the characters generated by our method exhibit superior performance compared to those produced by alternative approaches, particularly in the context of anime reference images. More results and analysis are given in the supplementary material. This underlines our method's superiority over competing methodologies.

In addition to automatic evaluation, we also conducted a user study to further assess these methods. During each round of testing, participants were asked to select their preferred image according to a specific metric from among four methods. For more detailed information, please refer to the supplementary material. 
The results demonstrate that our methods received the majority of preferences across all metrics, thus surpassing other methods. This finding is consistent with the outcomes of the automatic evaluations.

\begin{table}
\centering
\begin{tabular}{@{}lccc@{}}
\toprule
Method & CLIP-I $\uparrow$ & CLIP-T$\uparrow$ & Face Sim.$\uparrow$    \\
\midrule
IP-Adapter~\cite{ye2023ip} & 81.31 & 20.22 & 0.39 \\
FastComposer~\cite{xiao2024fastcomposer} & 73.98 & 21.43 & 0.32\\
StoryMaker~\cite{zhou2024storymaker} & 84.24 &  20.45 & \textbf{0.54}\\
Ours & \textbf{85.49} & \textbf{21.76} & 0.53 \\
\bottomrule
\end{tabular}
\caption{Comparison to other methods.}
\label{tab:comp_clip_score}
\end{table}


\subsection{Analysis and Ablation Study}

\begin{figure}[ht!]
    \centering
    \begin{subfigure}[b]{0.21\linewidth}
        \centering
        \includegraphics[width=0.95\linewidth]{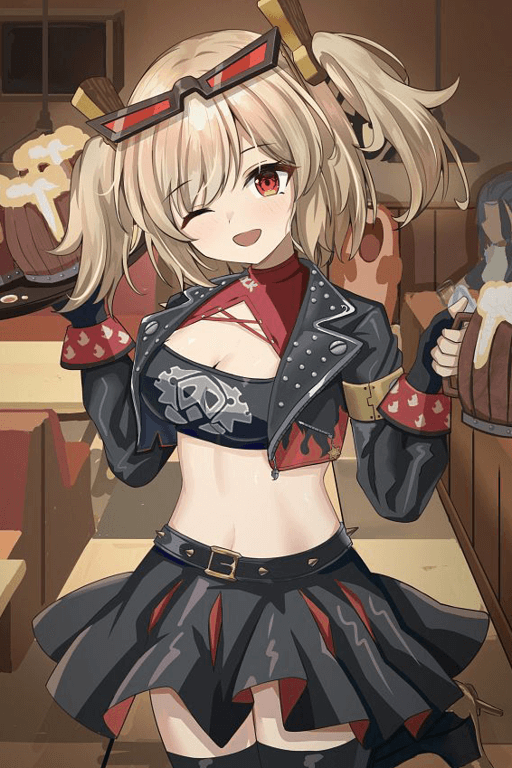}
        \caption{}
    \end{subfigure}
    ~ 
    \begin{subfigure}[b]{0.23\linewidth}
        \centering
        \includegraphics[width=0.97\linewidth]{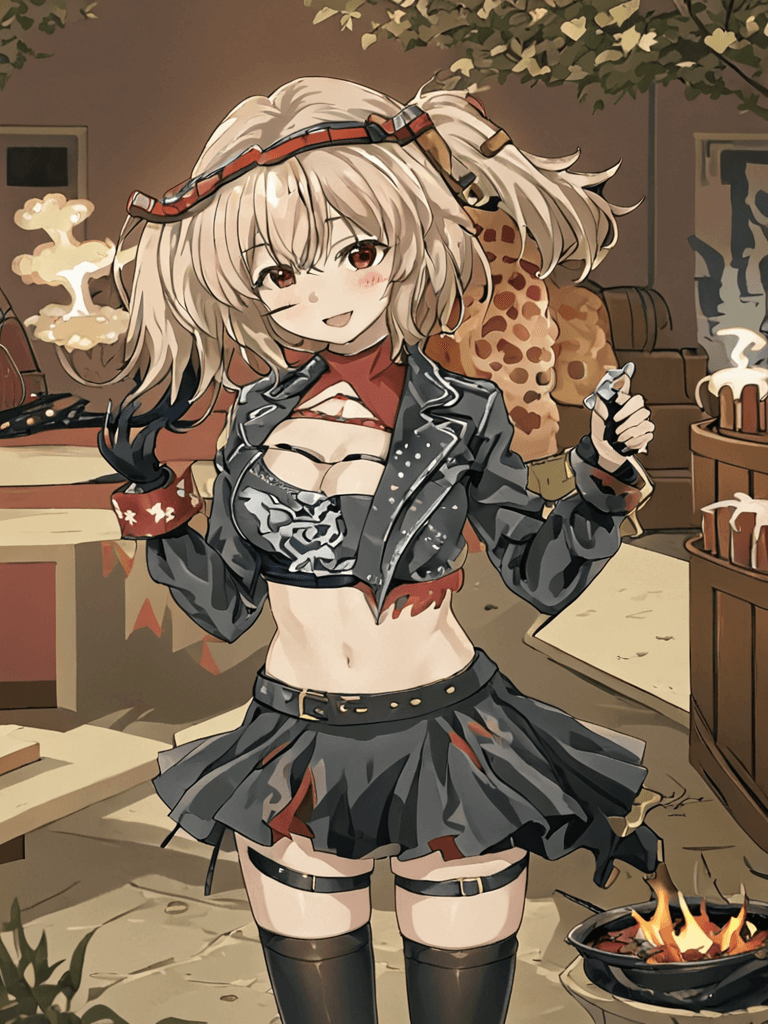}
        \caption{}
    \end{subfigure}
    ~
    \begin{subfigure}[b]{0.23\linewidth}
        \centering
        \includegraphics[width=0.97\linewidth]{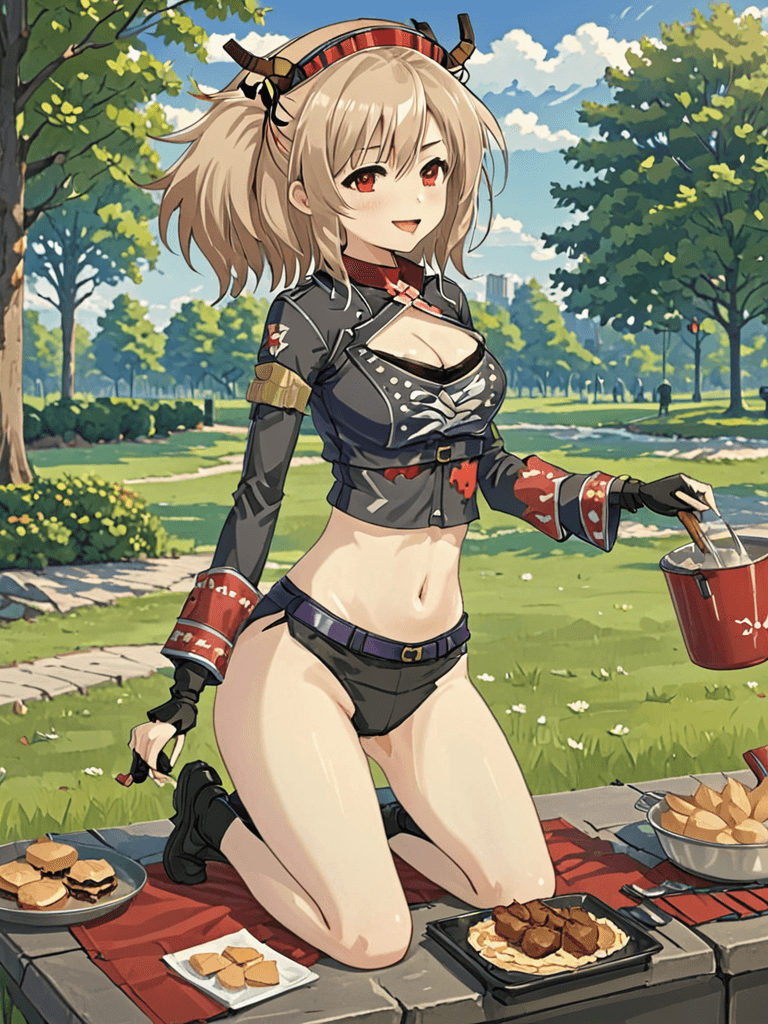}
        \caption{}
    \end{subfigure}
    ~
    \begin{subfigure}[b]{0.23\linewidth}
        \centering
        \includegraphics[width=0.97\linewidth]{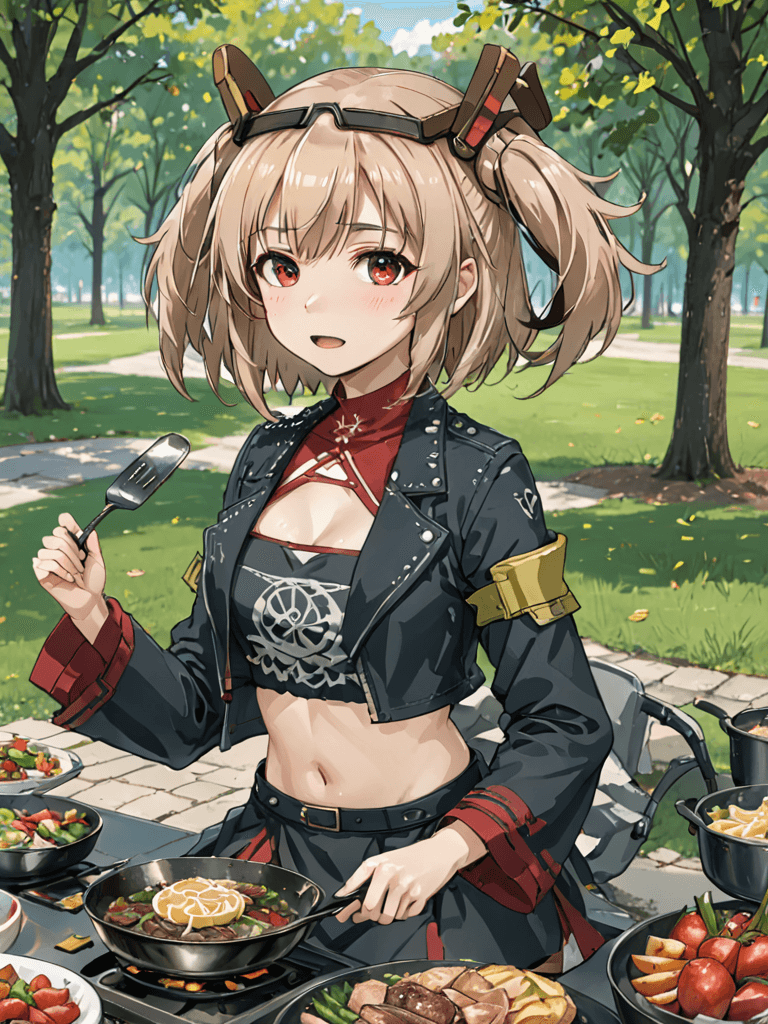}
        \caption{}
    \end{subfigure}
    \caption{Visual comparison of different training strategies. Prompt: \textit{1girl, cooking the meal, in the park}. (a) Reference Image; (b) Unpaired one-stage; (c) Paired one-stage; (d) Ours.}
   \label{fig:one_stage_ab1}
\vspace{-3mm}
\end{figure}

\begin{table}[t]
\centering
\begin{tabular}{@{}lccc@{}}
\toprule
Method & CLIP-I $\uparrow$ & CLIP-T$\uparrow$ & Face Sim.$\uparrow$   \\
\midrule
unpaired-one-stage  & 89.62 & 16.08 & 0.52\\
paired-one-stage    & 80.00 & 21.99 & 0.41\\
two-stage (ours)    & 85.49 &  21.76 & 0.53\\
\bottomrule
\end{tabular}
\caption{Comparison to one-stage methods.}
\label{tab:comp_diff_stage}
\end{table}

\subsubsection{Comparison to One-stage}

\begin{figure}[ht]
  \centering
   \includegraphics[width=0.9\linewidth]{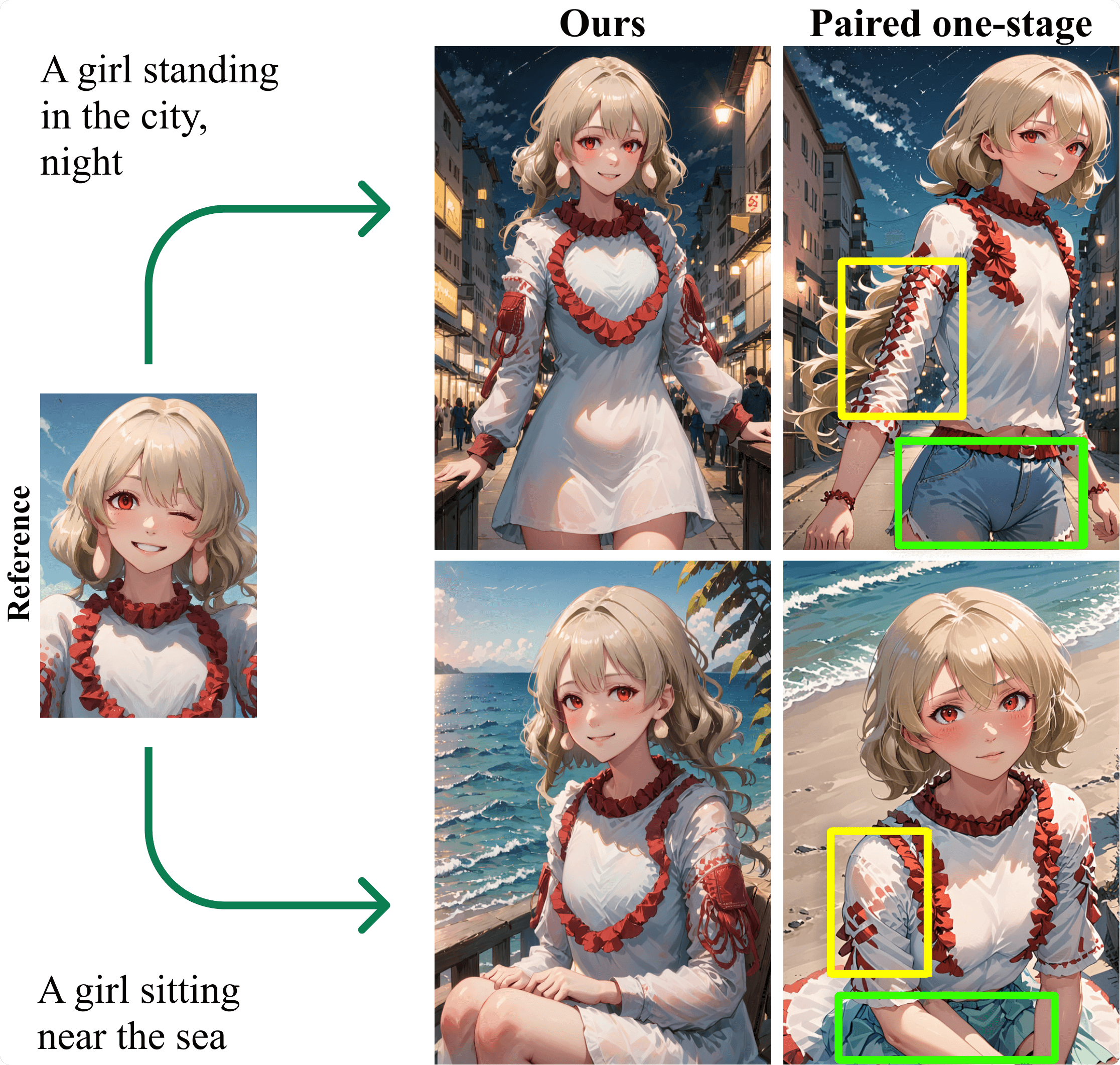}
   \caption{Consistency between two prompts. The inconsistent part is highlighted by rectangle.}
   \label{fig:one_stage_ab2}
\end{figure} 

We compare our two-stage personalization model against two one-stage models, each employing distinct training strategies: unpaired and paired. All three compared models share the same model architecture and are initially given the same training dataset, $\mathcal{D}$, comprising 300,000 images. The unpaired one-stage model is trained directly on $\mathcal{D}$ using the loss function described in Equation ~\eqref{eq:diff}. Our model first performs standardization on $\mathcal{D}$ to get 300,000 pairs of (standardized reference image, target image) and then trains on it. The paired one-stage model is trained on paired data. However, these pairs are generated by applying our two-stage model to dataset $\mathcal{D}$, resulting in 300,000 pairs of images. The key distinction here is that the reference image of paired one-stage is not standardized and only one stage inference is required.

The results are presented in Table \ref{tab:comp_diff_stage} and Figure \ref{fig:one_stage_ab1}. It is observed that the unpaired one-stage model exhibits a replication issue, where major contents of the reference image are duplicated in the output. This results in a high CLIP-I score of $89.62$ and a Face Sim. score of $0.52$ but a significantly low CLIP-T score of $16.08$, indicating that the text prompts are not accurately represented in the output images. As illustrated in Figure \ref{fig:one_stage_ab1}(b), the background and action do not closely align with the prompt. This highlights the limitations associated with unpaired training. Further experimental analysis is provided in the supplementary materials, where we demonstrate that unpaired training may also suffer from low appearance consistency while exhibiting high text controllability.
On the other hand, the remaining two methods, which utilize paired data, exhibit higher scores of $21.99$ and $21.76$ on the CLIP-T metric. The visualization outcomes further reveal that these methods can accurately respond to the text prompts. \textbf{This demonstrates the superiority of paired training over unpaired training}.

Reducing our model from a two-stage to a paired one-stage configuration results in a significant decrease of $5.49$ in the CLIP-I score. As evident from Figure \ref{fig:one_stage_ab1}(c) and (d), the lower body clothing of the character does not match the reference image in the paired one-stage model, whereas ours maintains consistency. \textbf{This highlights that utilizing standardized references enhances the consistency between the reference image and the output}.

We demonstrate that \textbf{standardization improves appearance consistency across serial images generated from different text prompts}. This form of appearance consistency among multiple outputs is quantified by averaging all pairwise CLIP-I (AP-CLIP-I) scores. 
 Our model achieves an AP-CLIP-I score of $83.17$. In contrast, the score decreases to $77.74$ for the paired one-stage model without standardization. The qualitative results presented in Figure \ref{fig:one_stage_ab2} further underline the importance of standardization. As illustrated, when the reference image displays only a character’s head, the absence of standardization can result in variations in the body’s appearance across different generative processes. Standardization addresses this issue by pre-generating the body in the standardized reference image, thereby ensuring a consistent appearance across various text prompts.

\begin{table}
\centering
\begin{tabular}{@{}lcc@{}}
\toprule
\textbf{}      & PSNR$\uparrow$ & SSIM$\uparrow$ \\
\midrule
baseline & 12.60          & 0.5820          \\
baseline+FBDM & 12.79          & 0.5898         \\
baseline+RPIM+FBDM& \textbf{12.84} & \textbf{0.5916} \\
\bottomrule
\end{tabular}
\caption{Ablation study of the standardization modules.}
\label{tab:arc_dis}
\end{table}

\begin{figure}[t]
  \centering
   \includegraphics[width=1\linewidth]{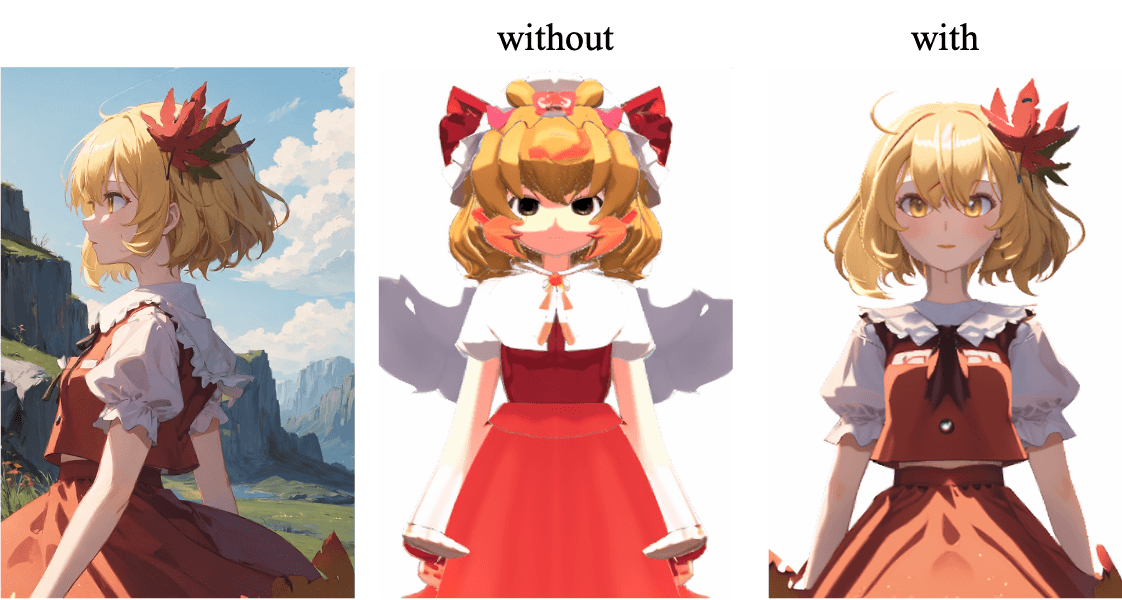}
   \caption{Standardization results with and without proposed FBDM\&RPIM. Leftmost is the input.}
   \label{fig:exp-ab}
\end{figure}

\subsubsection{The Architecture of Standardization Model}
\label{sec:architecture}
In this part, we verify the effectiveness of the proposed FBDM and RPIM modules. We first conduct experiments on our synthetic data, which is divided into a training set (80\%) and a testing set (20\%) based on the character ID. Table \ref{tab:arc_dis} presents the PSNR~\cite{hore2010image} and SSIM~\cite{wang2004image} for predictions made by the standardization model, both with and without the integration of the proposed modules. The results illustrate the superiority of the proposed modules.
Qualitative results, depicted in Figure \ref{fig:exp-ab}, demonstrate that the incorporation of the proposed modules enhances the standardization outcomes, yielding improved consistency.	

Next, we demonstrate the impact of the proposed modules on the final results of the two-stage generation. As shown in Table \ref{tab:comp_dis_model}, the inclusion of these modules improves all evaluated metrics, emphasizing their importance.

Finally, we discuss the rationality of overall architecture choice made for our standardization model. Given that the standardization task fundamentally pertains to human image animation, we compare our architecture against leading human image animation models using the benchmark dataset TikTok~\cite{wang2024disco}. 
No additional training data was utilized to ensure a fair comparison. Finally, our method achieves an FVD of $149.95$ and FID-VID of $14.75$, outperforming state-of-the-art methods and justifying our architectural choices. More quantitative results are provided in the supplementary material.

\begin{table}
\centering
\begin{tabular}{@{}lccc@{}}
\toprule
Module & CLIP-I $\uparrow$ & CLIP-T$\uparrow$  & AP-CLIP-I$\uparrow$   \\
\midrule
without        & 84.91 & 21.71  &82.86\\
with  & \textbf{85.16} & \textbf{21.85} & \textbf{83.15}\\
\bottomrule
\end{tabular}
\caption{Two-stage generation results with and without proposed FBDM\&RPIM in the standardization model.}
\label{tab:comp_dis_model}
\end{table}
\section{Conclusion}
Our work focuses on balancing text controllability with appearance consistency. By developing SerialGen which employs a serial generation method, we have empirically shown that it is possible to achieve high appearance consistency and adequate responsiveness to diverse text prompts. 
This work not only contributes valuable insights to the community but also showcases the practical implications of our approach in applications requiring high fidelity and personalized output, such as in comic story generation.
{
    \small
    \bibliographystyle{ieeenat_fullname}
    \bibliography{main}
}
\clearpage
\setcounter{page}{1}
\maketitlesupplementary

\begin{strip}
\centering
\includegraphics[width=0.95\textwidth]{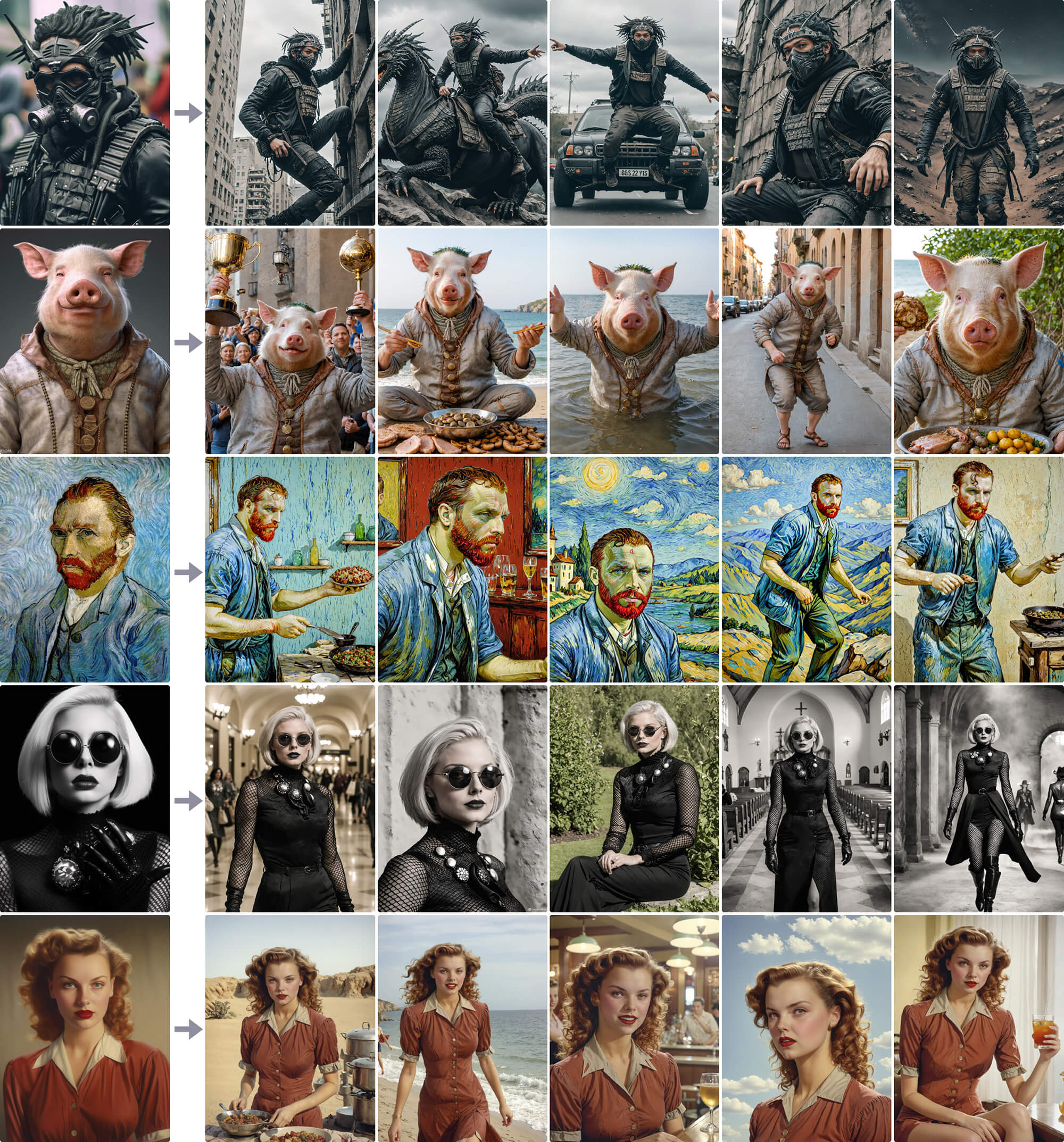}
\captionof{figure}{More serial images generated by SerialGen, showcasing its outstanding ability to maintain whole-body appearance consistency across different types of characters, including non-human subjects.}
\label{fig:supphomepage}
\end{strip}

\clearpage
\begin{strip}
\centering
\includegraphics[width=0.95\textwidth]{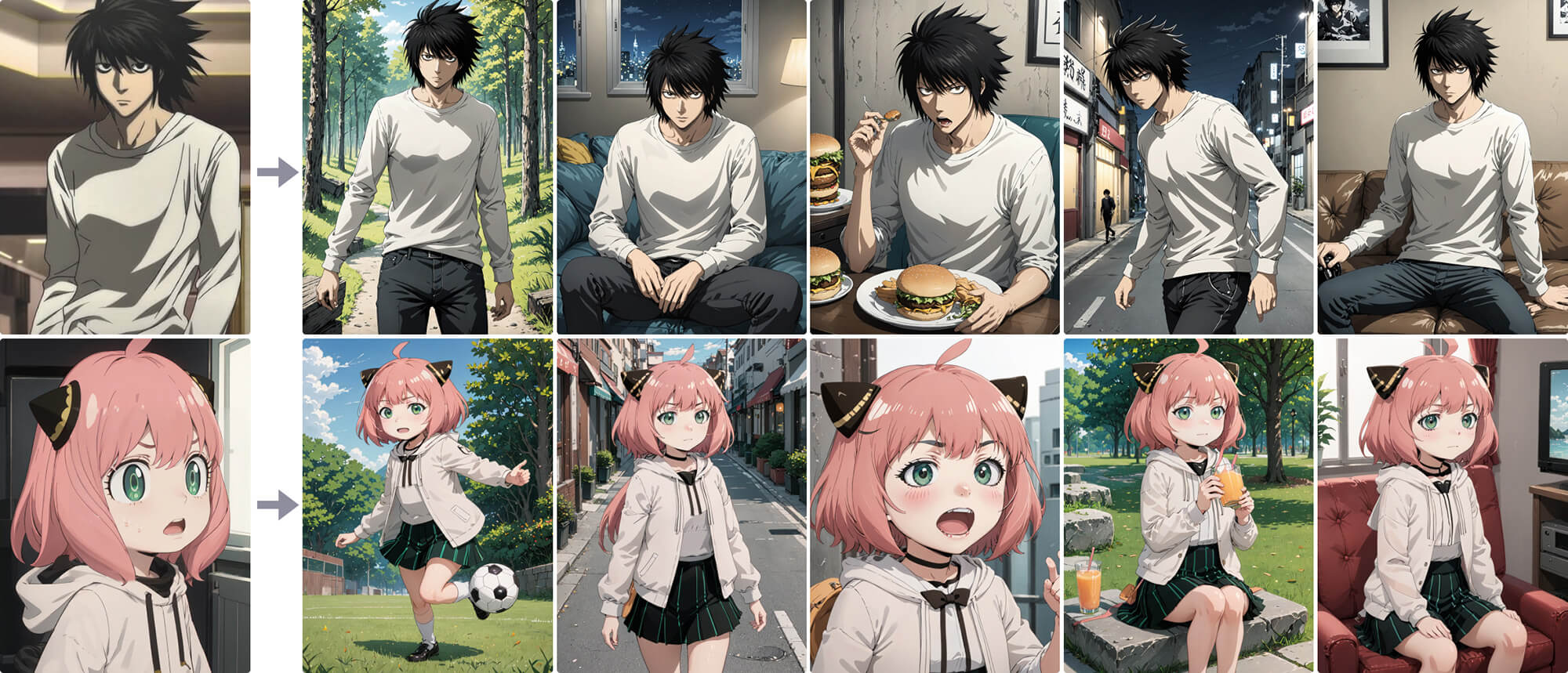}
\captionof{figure}{Extension of Figure~\ref{fig:supphomepage}.}
\end{strip}

\section{Details of the Reference Pose Injection Module}
\begin{figure}[h]
  \centering
   \includegraphics[width=1\linewidth]{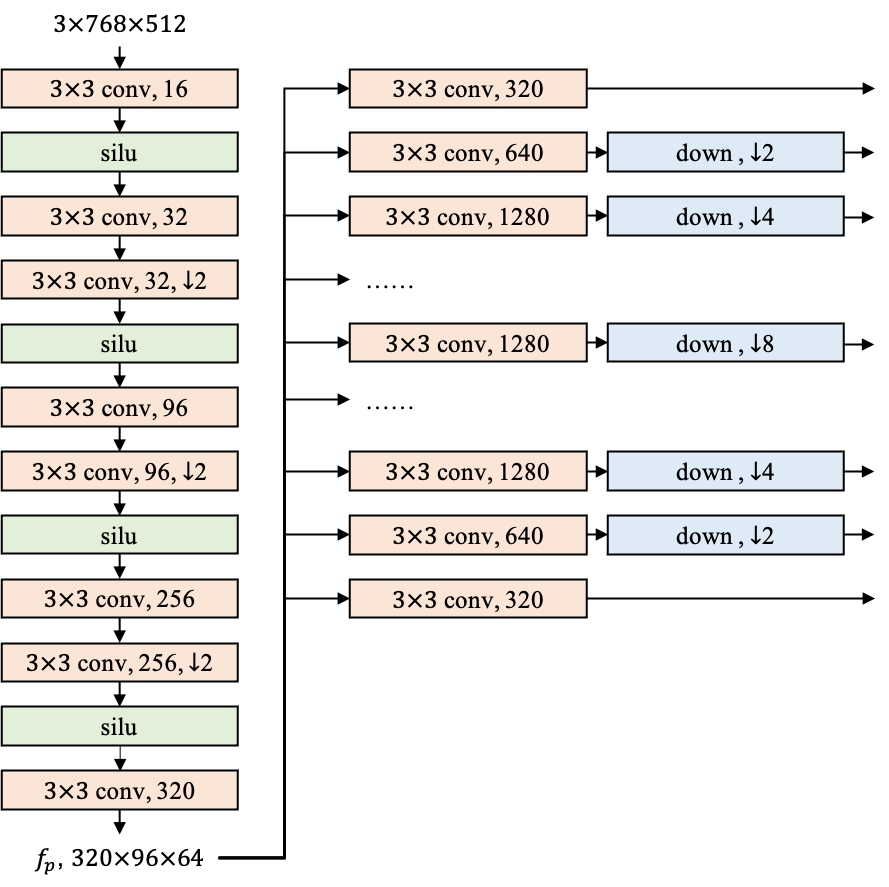}
   \caption{Details of the Reference Pose Injection Module.}
   \label{fig:rpim}
\end{figure}

We utilize a light convolutional network to extract pose feature maps from pose images. The architectural setup is depicted in Figure \ref{fig:rpim}, where $3 \times 3$ conv, $32$, $\downarrow 2$ indicates a convolutional layer with a kernel size of $3 \times 3$, a channel number of 32, and a stride size of 2. The term silu refers to a SiLU activation layer. The network processes the input through several convolutional stages with channel counts of $16, 32, 96$, $256$, and $320$, progressively reducing the spatial resolution by a factor of $8$. Before being added to the $\tilde{\mathbf{f}_r}$ feature in each self-attention module, an additional convolutional layer is introduced, followed by interpolation-based downsampling to align the dimensions of pose feature with the $\tilde{\mathbf{f}_r}$ feature.

\section{Impact of 3D Style Bias}
\begin{figure*}[h]
  \centering
   \includegraphics[width=\linewidth]{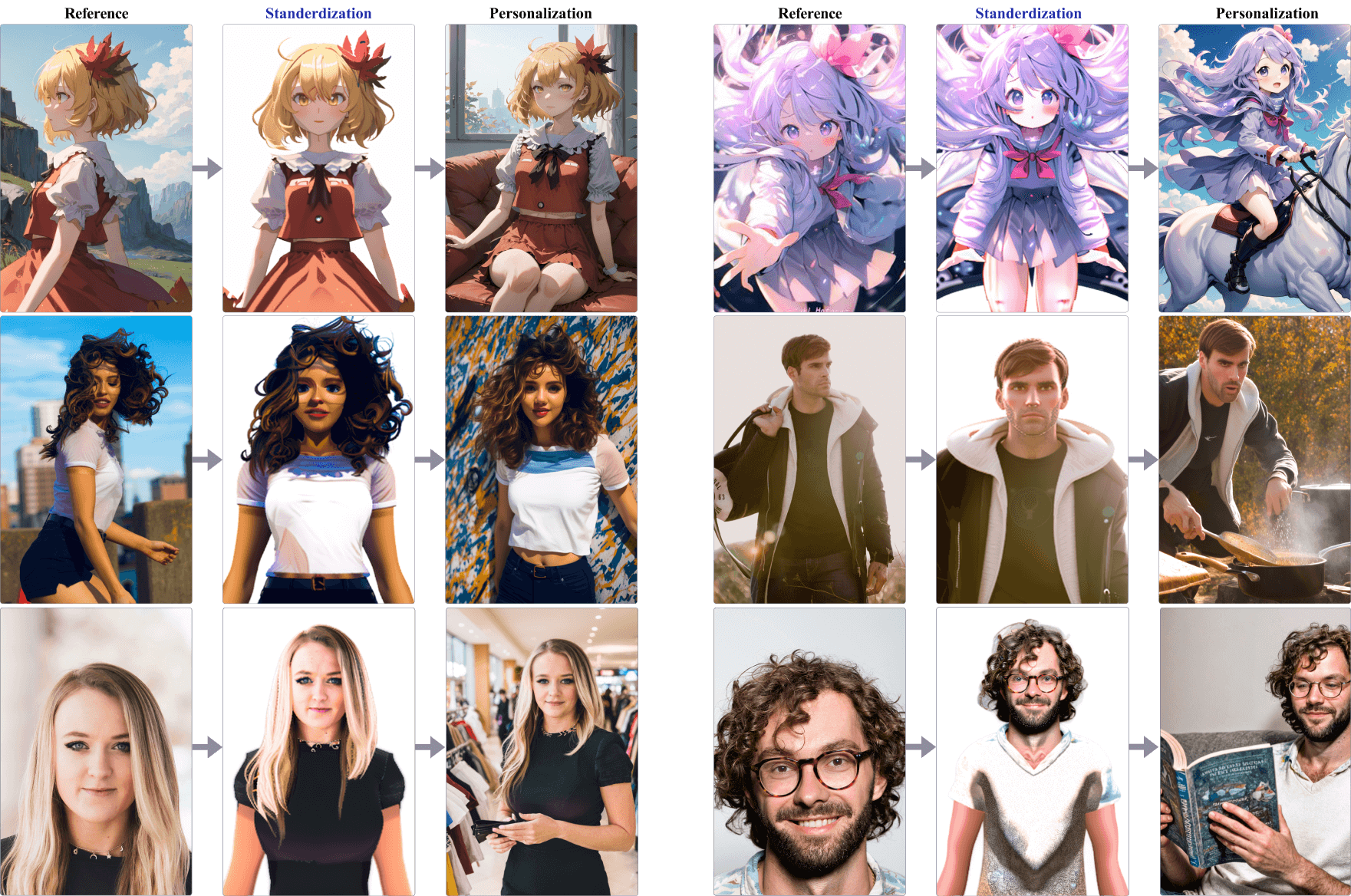}
   \caption{The standardization introduces a slight 3D style bias, particularly evident in head-only inputs (last row), resulting in clothing with a 3D appearance. This bias is effectively mitigated during the personalization stage.}
   \label{fig:supp_3d}
\end{figure*}

As depicted in the second paragraph of Section \ref{sec:personalization}, we demonstrate the impact of 3D style bias introduced by the synthetic data.
As shown in Figure \ref{fig:supp_3d}, the standardization stage introduces a slight 3D style bias when standardizing images. This bias is effectively mitigated during the personalization stage.
Specifically, as shown in the last row of Figure \ref{fig:supp_3d}, given a head-only image, the standardization stage generates clothing with a noticeable 3D style. However, the personalization stage subsequently recovers realistic clothing appearances.

\section{More Comparison Results and Analysis} 

\begin{figure*}[htb]
  \centering
   \includegraphics[width=\linewidth]{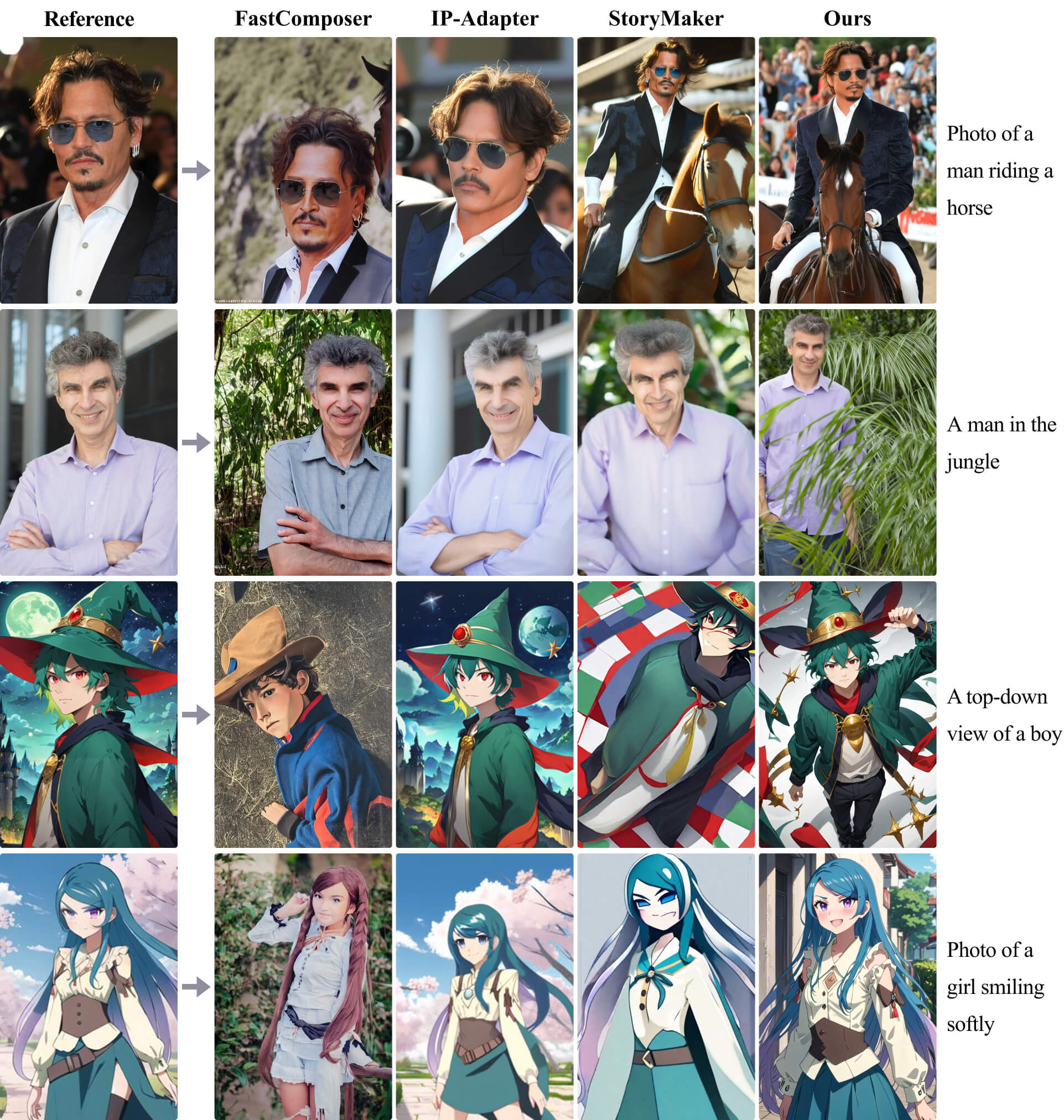}
   \caption{More comparison with other methods.}
   \label{fig:supp_more_comp}
\end{figure*}

This part gives supplementary comparisons and analysis in Section \ref{sec:quantitative}.
We give more comparison results with FastComposer, IP-Adapter and StoryMaker. As shown in the Figure~\ref{fig:supp_more_comp}, we selected four different characters for analysis, which include two anime characters and two real-life humans. The evaluation prompts are categorized into four descriptive types: action, background, viewpoint, and expression, arranged from the first row to the fourth row, respectively.  In the first row, both StoryMaker and our method successfully alter the character’s action, with our method maintaining a more accurate hairstyle.  In the second row, both FastComposer and our method produce images featuring a clear jungle background; however, FastComposer does not accurately depict the character’s clothes.  In the third and fourth rows, while IP-Adapter manages to capture the anime character’s appearance, it struggles to modify the viewpoint and expression, a limitation attributed to its unpaired one-stage training strategy. Conversely, our method effectively generates images that match the descriptions of expressions and viewpoints accurately. Our approach demonstrates superior performance in maintaining appearance consistency and textual controllability compared to other leading-edge methods.  
\begin{table}
\centering
\begin{tabular}{@{}lccc@{}}
\toprule
Method & WAC & TC & VA \\
\midrule
IP-Adapter~\cite{ye2023ip} & 20.00\% & 4.33\% & 5.67\% \\
FastComposer~\cite{xiao2024fastcomposer} & 4.67\% & 3.67\%& 0.67\%\\
StoryMaker~\cite{zhou2024storymaker}   & 20.33 \% & 37.67\% & 22.67\%\\
Ours  & \textbf{55.00\%} & \textbf{54.33\%} & \textbf{71.00\%} \\
\bottomrule
\end{tabular}
\caption{User preference in personalized image generation, evaluated across three criteria: whole-body appearance consistency (WAC), text controllability (TC), and visual appeal (VA).}
\label{tab:user_study}
\end{table}
\section{User Study}

As shown in Table \ref{tab:user_study}, we design three criteria for comparison, where each criterion receives 600 valid votes (30 participant $\times$ 20 text-image pairs). The detailed questions are as follows: 1) Whole-body Appearance Consistency: Which method best preserves the input character's whole-body appearance? 2) Text Controllability: Which method generates images that best align with the input text prompt? 3) Visual Appeal: Which method produces the most visually appealing image? To ensure objectivity, the names of all methods are anonymized, and the methods are presented in a randomized order for each question.

\section{Limitations of Unpaired Training}

In these experiments, we train models on unpaired image data, using identical images as both reference and target. For the reference encoder, we employ IP-Adapter ~\cite{ye2023ip}, while SDXL is utilized as the diffusion model. The feature size extracted from the reference image can be adjusted by modifying a setup parameter in IP-Adapter, known as the number of token features. An increase in the number of token features corresponds to a more powerful reference encoder. 

\begin{figure*}[ht!]
\centering
\includegraphics[width=\linewidth]{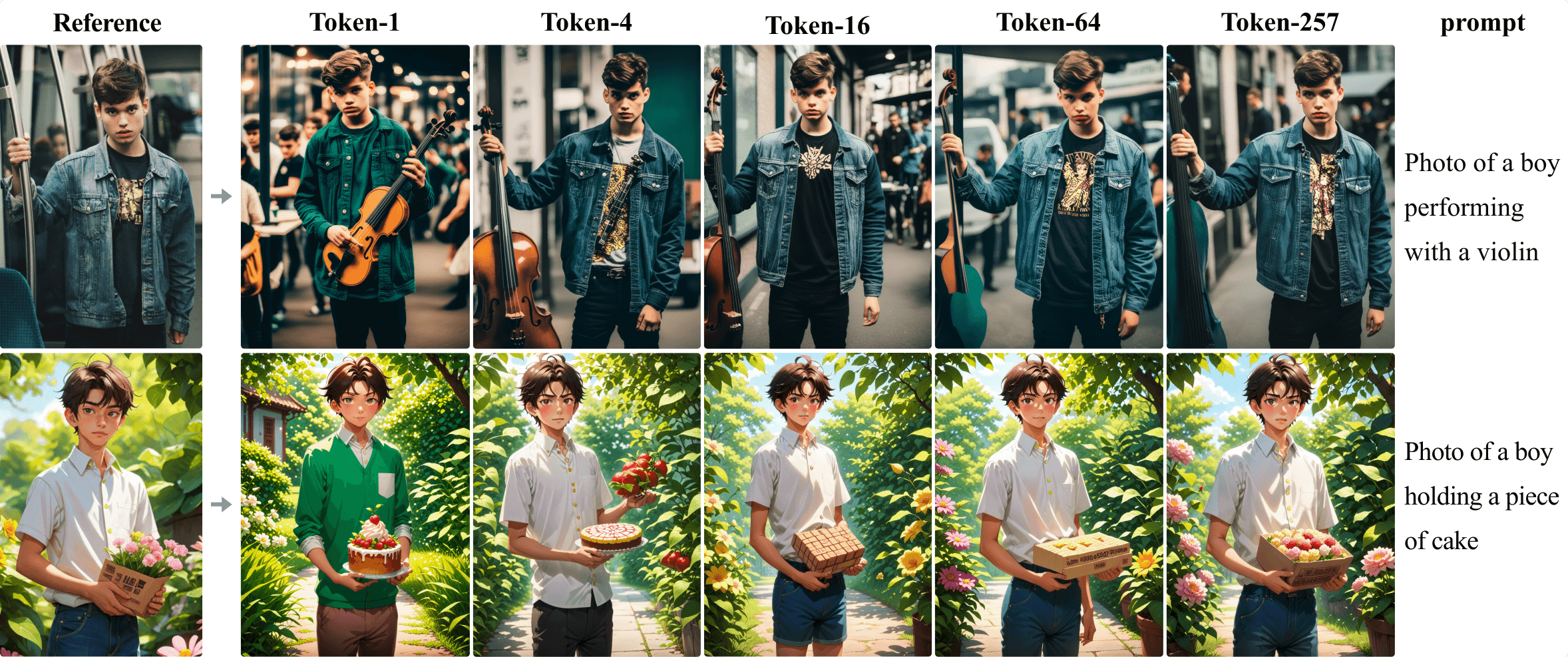}
\caption{Visual comparison of different numbers of token features. Leftmost is the reference image. Token-$i$ indicates the model trained with a token feature number of $i$.}
\label{fig:token_vis}
\end{figure*}

\begin{figure}[ht]
  \centering
   \includegraphics[width=1\linewidth]{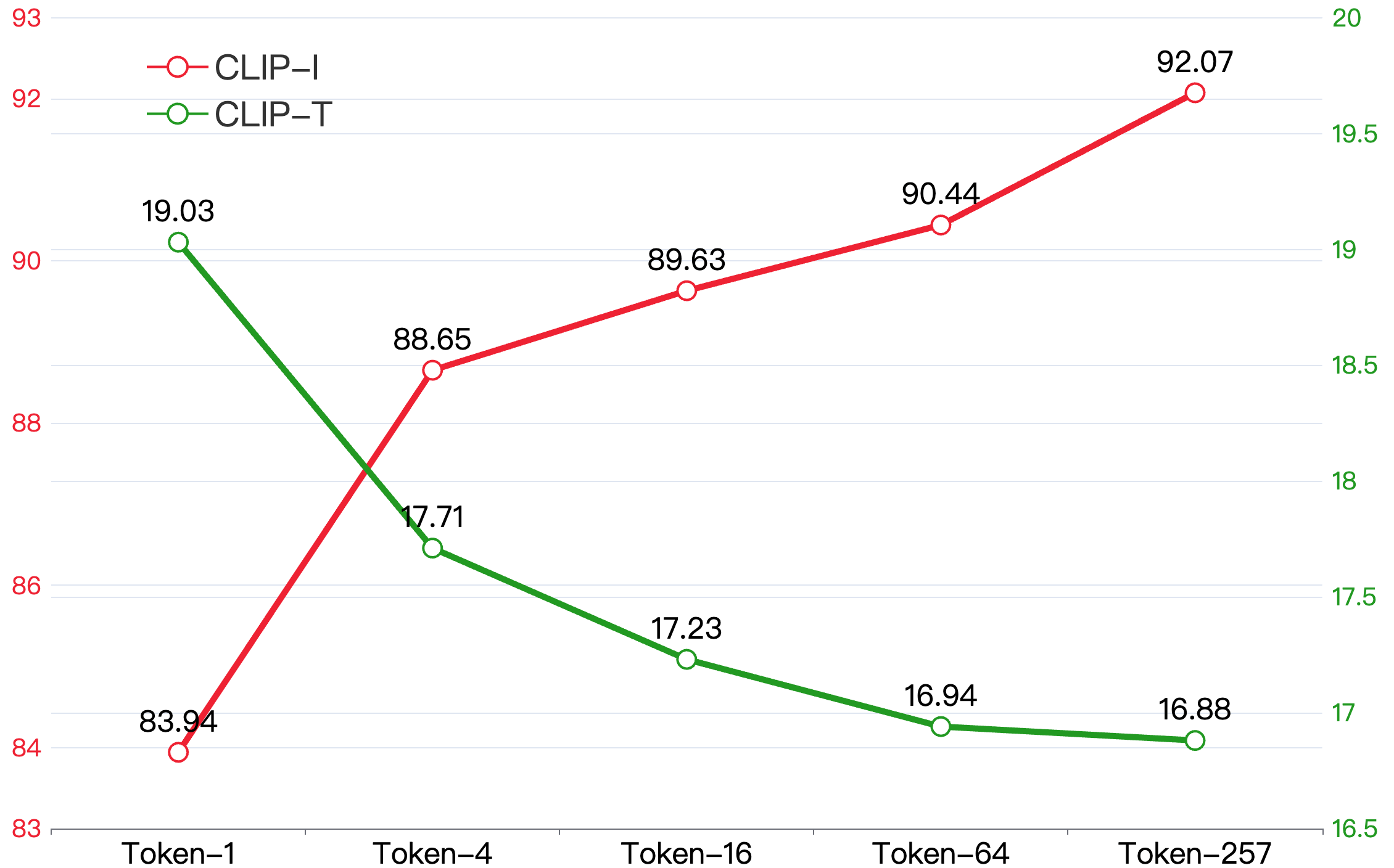}
   \caption{Comparison of different numbers of token features .}
   \label{fig:token_comp}
\end{figure}

We systematically train a series of IP-Adapter reference encoders, including both powerful and weak configurations, by varying the number of token features. Figure~\ref{fig:token_vis} presents some qualitative results with varying numbers of tokens. From these results, it is evident that unpaired training struggles to meet the objectives of personalized generation tasks: the models either compromise text controllability to maintain high appearance consistency or sacrifice appearance consistency to enhance text controllability. In the settings of powerful encoders—those equipped with a larger number of tokens—the models can easily replicate the reference image, achieving high appearance consistency but showing inadequate responsiveness to text prompts. Conversely, in the settings of weak encoders, there is a better alignment with text prompts, albeit at the expense of compromised appearance consistency. The quantitative results depicted in Figure~\ref{fig:token_comp} also align with these visual observations. As the number of tokens increases, indicating more powerful encoders, there is an observed rise in the CLIP-I score, from $83.94$ to $92.07$, while the CLIP-T score decreases, moving from $19.03$ to $16.88$.

\section{Comparison to Human Image Animation Models}

\begin{table}[!t]
\centering
\begin{tabular}{lcc}
\toprule
Method & FVD$\downarrow$ & FID-VID$\downarrow$ \\
\midrule
DisCo~\cite{wang2024disco} & 292.8 & 59.9 \\
MagicPose~\cite{chang2023magicpose} & - & 46.3 \\
MagicAnimate~\cite{xu2024magicanimate} & 179.07 & 21.75 \\
Animate Anyone~\cite{hu2024animate} & 171.9 & - \\
Champ~\cite{zhu2024champ} & 160.82 & 21.07 \\
TCAN~\cite{kim2024tcan} & 154.84 & 19.42 \\
Ours & \textbf{149.95} & \textbf{14.75} \\
\bottomrule
\end{tabular}
\caption{Quantitative comparison on TikTok dataset.}
\label{tab:tiktok}
\end{table}

As discussed in Section \ref{sec:architecture}, we compare the architecture of our standardization model with other leading human image animation models, including DisCo~\cite{wang2024disco}, MagicPose~\cite{chang2023magicpose}, MagicAnimate~\cite{xu2024magicanimate}, Animate Anyone~\cite{hu2024animate}, Champ~\cite{zhu2024champ}, and TCAN~\cite{kim2024tcan}. Experiments are conducted using the benchmark dataset TikTok~\cite{wang2024disco}. No additional training data was utilized to ensure a fair comparison. To enable training on video datasets, a temporal layer~\cite{hu2024animate} is incorporated into the architecture described in Section \ref{sec:standardization}. The results presented in Table \ref{tab:tiktok} demonstrate that our method significantly outperforms existing state-of-the-art approaches, achieving superior performance in both FVD and FID-VID metrics. These results justify our architectural choices.

\section{Ablation Study on Identity Loss}

\begin{table}
\centering
\begin{tabular}{@{}lcc@{}}
\toprule
Module & CLIP-I $\uparrow$ & Face Sim.$\uparrow$   \\
\midrule
after standardization & 89.47 & 0.69 \\
after personalization & 85.49 & 0.53 \\
\bottomrule
\end{tabular}
\caption{Ablation study on identity loss at each stage.}
\label{tab:ab_idloss}
\end{table}
We conduct an ablation study to evaluate the impact of each stage on identity preservation. As shown in Table \ref{tab:ab_idloss}, after the standardization stage, CLIP-I is 89.47, and Face Sim. is 0.69. Following the personalization stage, CLIP-I decreases to 85.49, while Face Sim. drops to 0.53. These results indicate that identity consistency remains relatively high after the standardization stage.

\section{More Quantitative Comparisons}

We also made a quantitative comparison between our method and the recent face-oriented approach LCM-Lookahead~\cite{gal2024lcm}, which achieved a Face Sim. score of 0.46, CLIP-I score of 74.56, and CLIP-T score of 24.63 on the test dataset. Our method outperforms LCM-Lookahead in both CLIP-I and Face Sim. metrics, with only a disadvantage in CLIP-T. Notably, LCM-Lookahead achieves good text controllability at the cost of consistency.


\end{document}